\documentclass[runningheads]{llncs}

\usepackage{epsfig}
\usepackage{calc}
\usepackage{pslatex}
\usepackage{graphicx} % for pdf, bitmapped graphics files
\usepackage{mathptmx} % assumes new font selection scheme installed
\usepackage{newtxtext,newtxmath}
\usepackage{wrapfig}
\usepackage{adjustbox}
\usepackage[hidelinks]{hyperref}
\begin{document}
\title{Human-Likeness Indicator for Robot Posture Control and Balance}
%\title{Contribution Title\thanks{Supported by organization x.}}
\author{Vittorio Lippi\inst{1,2}\orcidID{0000-0001-5520-8974} \and Christoph Maurer\inst{2}\orcidID{0000-0001-9050-279X} \and Thomas Mergner\inst{2}\orcidID{0000-0001-7231-164X}}
\authorrunning{Lippi  et al.}
\institute{Institute of Digitalization in Medicine, Faculty of Medicine and Medical Center - University of Freiburg, Freiburg, Germany \and Clinic of Neurology and Neurophysiology, Medical Centre-University of Freiburg, Faculty of Medicine, University of Freiburg, Breisacher Stra{\ss}e 64, 79106, Freiburg im Breisgau \\
\email{{\{vittorio.lippi, christoph.maurer, thomas.mergner\}@uniklinik-freiburg.de}}}
\maketitle              % typeset the header of the contribution
\begin{abstract}
Similarly to humans, humanoid robots require posture control and balance to walk and interact with the environment. In this work posture control in perturbed conditions is evaluated as a performance test for humanoid control. A specific performance indicator is proposed: the score is based on the comparison between the body sway of the tested humanoid standing on a moving surface and the sway produced by healthy subjects performing the same experiment. This approach is here oriented to the evaluation of a human-likeness. The measure is tested using a humanoid robot in order to demonstrate a typical usage of the proposed evaluation scheme and an example of how to improve robot control on the basis of such a performance indicator score.

\end{abstract}
\section{INTRODUCTION}
\subsection{Overview}
The benchmarking of humanoid performance is gaining interest in the research community \cite{del2006benchmarks,behnke2006robot,conti2018people,torricelli2014benchmarking,torricelli2020benchmarking,Lippi2020,Lippi2019}.
 The performance of a humanoid is a complex issue that covers several aspects, e.g. sensor fusion, cognitive and motor functions, mechanics, and energy efficiency. In particular, a recent European project, EUROBENCH \cite{Anthony1,Anthony3,Antony2} is proposing to implement standard and repeatable experimental procedures to evaluate and compare the performance of different robots. This work describes one specific posture control performance indicator to be implemented within the project, evaluating human-likeness based on the similarity between human and robot responses to external disturbances \cite{Lippi2019,Lippi2020,robovis21}.

Posture control is here particularly relevant because falling is one of the typical reasons for failures for humanoids \cite{atkeson2015no,atkeson2018happened,guizzo2015hard}. This work focuses on posture control from the point of view of human-likeliness. A formal and unanimous definition of human-likeness is still missing, although the concept is relevant both for robotics and neuroscience \cite{torricelli2014benchmarking}. 
\begin{figure}[t!]
	\centering
		\includegraphics[width=1.00\columnwidth]{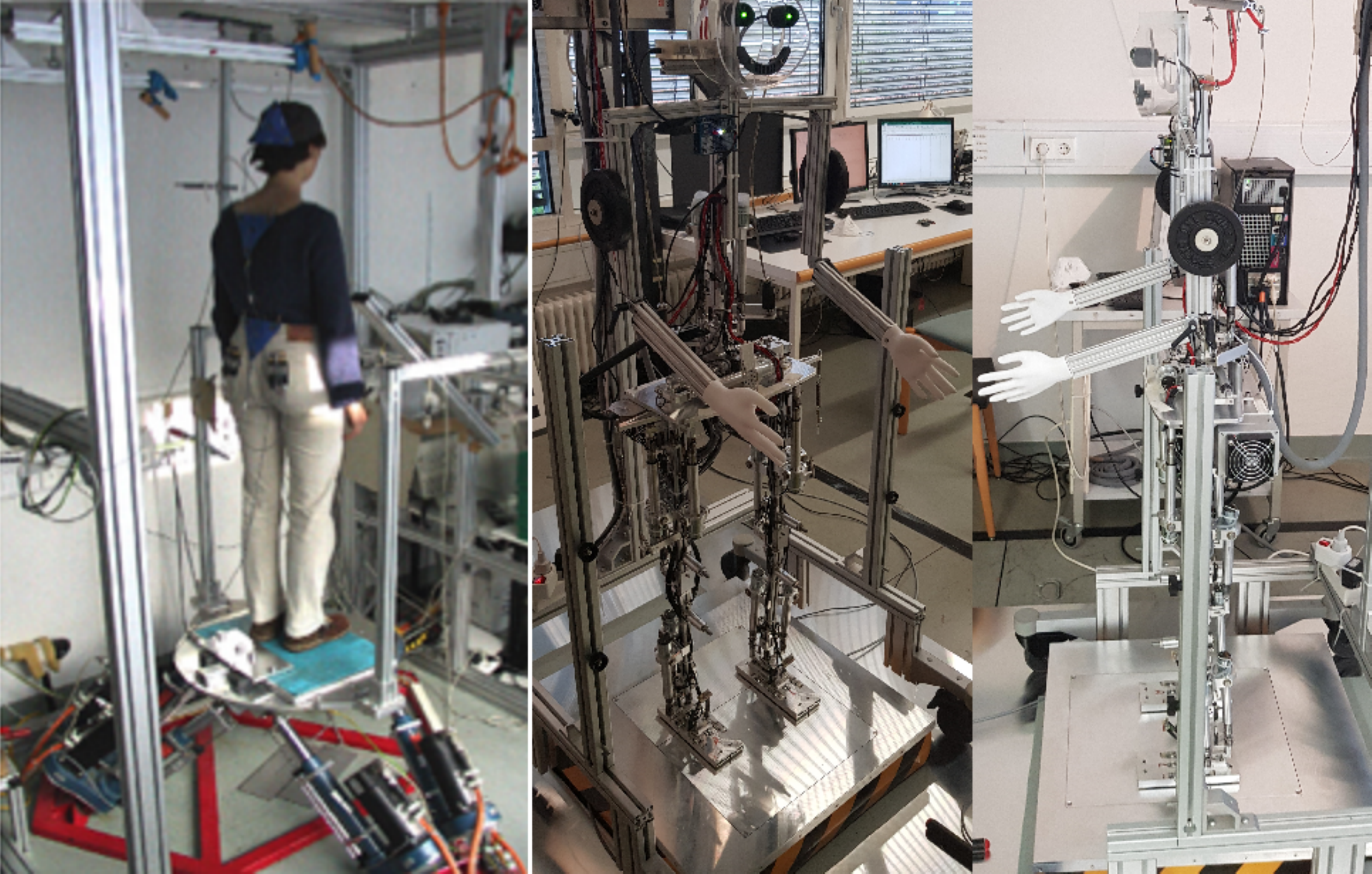}
	\caption{Posture control experiment with a moving support surface, a 6-Dof Stuart Platform. On the left, a human subject with active markers visible on the back (blue triangular plates) is shown. On the right, the \textit{Lucy} robot standing on a different platform that implements the same support surface tilt profile.}
	\label{fig:humanRobot}
\end{figure}
Notwithstanding the difficulty in defining human-like behavior, it is expected to be superior to the one of state-of-the-art robots \cite{nori2014whole}. Human-likeness is envisaged to be associated with some advantages such as mechanical compliance (also important for safety) and low energy consumption. Both of these features are typically associated with the low feedback loop gain typical for biological systems in face of the presence of neural delay. In robotics, the delay associated with the transport of signals is usually negligible, but the robustness in face of delays may be important in limiting the required computational resources needed to implement the real-time control \cite{ott2016good}. Often human-likeness is defined (1) in terms of perception from the point of view of human observers   \cite{von2013quantifying,oztop2005human,abel2020gender}, (2) identified with the presence of a specific feature, e.g. reproducing human trajectories \cite{kim2009stable} or exhibiting mechanical compliance \cite{colasanto2015bio}, or defined in relation to a set of tasks, e.g. in challenges as used in the DARPA challenge \cite{luo2014robust,dedonato2015human}. The measure proposed here is based on body kinematics with the aim to provide an indicator of human-likeness that is repeatable and objective.

The idea of testing robots in the experimental setup used for human posture control dates was inspired by the necessity of implementing and testing human posture control models in a ``real world'' set-up, including factors that are not usually well accounted for by simulations such as sensor and actuation noise \cite{G.Hettich2014,lippi2017human,Mergner2010,T.Mergner2009}. Once this approach was defined, posture control experiments provided a way to compare alternative control systems (bio-inspired or not) on the same robotic platform \cite{alexandrov2017human,ott2016good}. The general benchmarking issue represents a generalization and a formalization of such studies. In general, the mutual inspiration between humanoid robotics and neuroscience is well recognized and considered promising in both the robotics and neuroscience fields \cite{cheng2007cb}.

This article represents an extended version of \cite{robovis21}, a conference paper where the benchmarking principle was tested with simulated experiments. In the present work, the system is tested with a real humanoid robot.
In the remainder of the Introduction, we describe the proposed benchmarking approach summarizing the method previously detailed in \cite{robovis21}. In the Method section, the robotic platform, the control system, and the experimental testbed are described as well as the set-up used to produce the human data-set. The Results section will show the scores obtained with different settings in the control system. The implications of the proposed performance indicator and the limitations associated with human likeliness in humanoid posture control will be discussed in the conclusions.
 
\subsection{Description of the Approach}
The proposed benchmarking measure is based on a data-set of results from human experiments. The experimental data consist of the body sway of human body segments induced by an external stimulus, here specifically the tilt of the support surface in the body sagittal plane. The measured body sway is characterized as a \textit{Frequency Response Function}, FRF, i.e. an empirical transfer function between the stimulus characterized by a Pseudo Random Ternary Sequence (PRTS) profile \cite{peterka2002sensorimotor} and the response computed on specific frequency points. The data-set has been produced with healthy subjects. The comparison aims to assess the similarity in the balancing behavior between the robot and a set of average healthy subjects. Such measure does not make explicit assumptions about specific advantages of the human behavior, in contrast to the performance indexes proposed in other studies that identify some goal like minimum torque or energy consumption in robots \cite{icinco16}, or a specific problem in human subjects \cite{singh2016detection}.

Body sway profiles FRFs have been chosen as a basis for the benchmarking because such analysis has already been studied in several previous publications with human subjects. These provide a reference for comparison and tools for analysis. There are several reasons why the body sway induced by support surface tilt has been repeatedly used to study human posture control. As a repeatable stimulus, it can be used to formally characterize the behavior in terms of responses to a well-defined input. Furthermore, the tilting support surface requires the balancing mechanism to integrate vestibular input and proprioception (and vision, with eyes open), and hence it is well apt to study human sensor fusion mechanisms.\\

\subsection{Human Data-Set} A group of 38 young healthy subjects serves as a reference for the human-likeness criterion. The subjects were presented with a stimulus consisting of a tilt of the support surface in the body sagittal plane, while the recorded output was the body sway. The typical set-up is shown in Fig. \ref{fig:humanRobot}, the body sway tracking was performed using active markers (Optotrak 3020; Waterloo, ON, Canada), attached to subjects' hips and shoulders and the platform. A PC with custom software was used to generate the support surface tilt in the body-sagittal plane. The marker positions were recorded at 100 Hz using software written in LabView (National Instruments; Austin, TX, United States). The profile used for the stimulus is a pseudorandom ternary signal, PRTS \cite{peterka2002sensorimotor}. The peak-to-peak amplitude was set to $1^{\circ}$. The amplitude is rather small compared to what a healthy subject can withstand. Usually, in similar studies, the tilt may go up to 8$^{\circ}$  and more \cite{peterka2002sensorimotor,asslander2015visual}. This was motivated by the aim to provide a data-set that could be compared safely with elderly subjects and patients and, in the specific case of robotic benchmarking, can be used to characterize the behavior of the robot without the risk of making it fall (and potentially break). \label{dataset}%
\subsection{Performance indicator} 
\begin{figure}[htbp]
	\centering
		\includegraphics[width=1.00\textwidth]{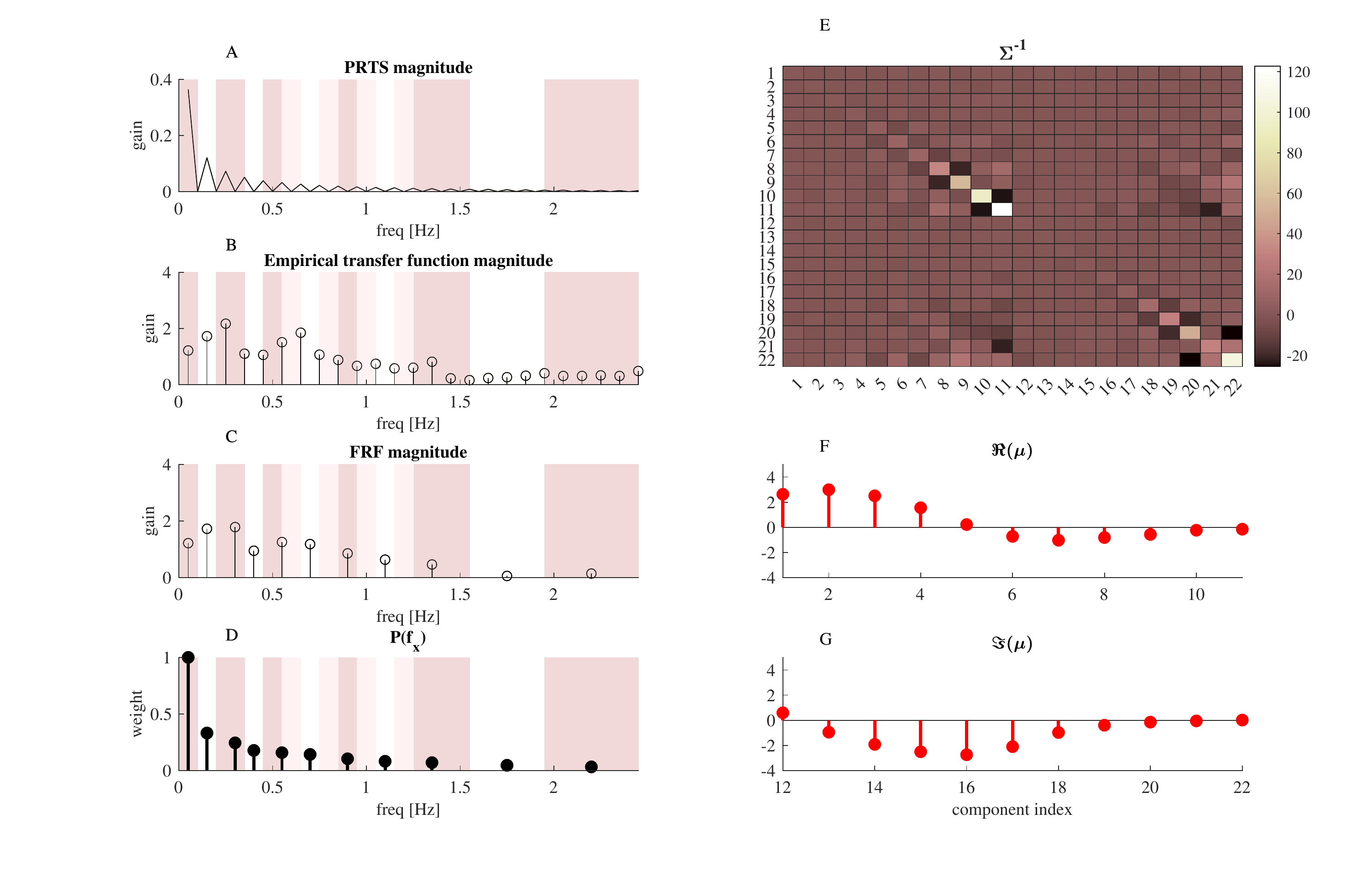}
	\caption{Input and parameters used in the definition of the performance indicator. \textbf{A} Magnitude of the spectrum of the input PRTS, with the typical ``comb'' profile alternating peaks to zeros. \textbf{B} Example of the empirical transfer function from a trial. \textbf{C} 11-component FRF associated with the transfer function in B. Note that, as the FRF is averaged in the complex domain the magnitude of the average shown in the plot is not the average of the magnitudes. The bands on the background show the frequency ranges over which the spectrum is averaged: white and pink represent ranges associated with groups of frequencies. The sets of frequencies are overlapping, with light-pink bands belonging to both the two contiguous groups, and a sample on the transition between two bands belongs to both the groups. \textbf{D} Weights representing the values of $P(f_x)$ in eq. \ref{distance}. \textbf{E} Inverse of $\Sigma$ from eq. \ref{distance}. Notice how the values on the diagonal corresponding to the highest frequencies are associated with high values (less variance in the dataset). This would make the metrics particularly sensitive to accidental changes in such components (e.g. due to noise). This is compensated by the weighting profiles in D, which associate almost zero weight to high-frequency components. \textbf{D} $\mathbf{\mu}$ from eq. \ref{Delta}. }
	\label{fig:Parameters}
\end{figure}

The performance indicator is here a measure of similarity with human behavior. The body sway profiles, i.e. the angular sway of the body COM with respect to the ankle joint, are used to characterize and compare the responses. The comparison is defined in terms of the norm of the difference between frequency response functions (FRFs). The PRTS power-spectrum has a profile with peaks at  $f_{peak}$ separated by ranges of frequencies with no power \cite{jilk2014contribution,peterka2002sensorimotor,icinco20}. %
The response is evaluated for a specific set of frequencies where the PRTS spectrum has peaks: $\mathbf{f_{peak}}$=[0.05, 0.15, 0.25, 0.35, 0.45, 0.55, 0.65, 0.75, 0.85, 0.95, 1.05, 1.15, 1.25, 1.35, 1.45, 1.55, 1.65, 1.75, 1.85, 1.95, 2.05, 2.15, 2.25, 2.35, 2.45] Hz. Such a discrete spectrum is then transformed into a vector of 11 components by averaging the FRF over neighboring frequencies as illustrated in Fig. \ref{fig:Parameters}:
	\begin{equation}
		f_{x(k)}=\frac{\sum_{i \in B_k} f_{peak(i)}}{N_k}
		\label{bands}
	\end{equation}
	where $k$ and $i$ are the indexes of the components of the frequency vectors, and $B_k$ is a set of $N_k$ frequencies averaged to obtain the $k^{th}$ sample. The $B_k$ are shown in Fig. \ref{fig:Parameters} as white and pink bands (notice that the bands are overlapping, light-pink has been used to indicate a zone belonging to two adjacent bands). Similarly the Fourier transform of the PRTS $P(f_{x})$ and the Fourier transform of the responses are averaged over the bands $B_k$ before computing the FRFs.   
The final representation of the FRF is a function of the 11 frequencies $f_{x}$ =[0.1, 0.3, 0.6, 0.8, 1.1, 1.4, 1.8, 2.2, 2.7, 3.5, 4.4]% 

In a previous description of the system in \cite{Lippi2020} the frequency vector had 16 components as proposed in other works using the PRTS, e.g. \cite{peterka2002sensorimotor,goodworth2018identifying,icinco20,lippi2020body}. In this work, a shorter version of the signal is used, which is considered safer and less fatiguing for human subjects, sometimes elderly patients, and convenient for robotics experimenters. Consequently, the discrete Fourier transform of the signal and the resulting FRFs have fewer components.

The choice of the frequencies in $B_k$ and their overlapping follows the method described in \cite{peterka2002sensorimotor}, but here is adapted to the 11 frequencies considered. The rationale behind such a choice was to get a representation with the frequencies equally spaced on a logarithmic scale, which is often used in posturography papers to present the FRF, with the overlapping providing a smoothing effect. 
The FRF is computed from the 11 components of the Fourier transform of the input $U$, and the output $Y$ as
\begin{equation}
	H=\frac{G_{UY}}{G_U}
\end{equation}
where $G_{UY}=U^* \odot Y$ and $G_U = U^* \odot U$ are empirical estimations of the cross power spectrum and the input power spectrum (``$\odot$'' is the  Hadamard, element-wise product). 
The peaks of the PRTS power-spectrum have larger values at lower frequencies \cite{jilk2014contribution}. This implies a better signal-to-noise ratio for the first components. A weighting vector $\mathbf{w}$ based on $P(f_{x})$ is then defined in a similar way to eq. \ref{bands}, but considering the power 
\begin{equation}
	w_k=\sqrt{\sum_{i \in B_k} ||P(f_{peak(i)})||^2}
	\label{weightsfactors}
\end{equation}.
The distance between two FRFs is defined and the norm of the difference weighted by the precision matrix, i.e. the inverse of the covariance matrix $\Sigma$, computed for the data-set of normal subjects. Before doing this, the FRF is expanded into a vector with the real and imaginary components as separated elements, i.e. 22 components. This together with the foretold weighting leads to the definition of the norm:
\begin{equation}
D=\sqrt{S \Delta^T \Sigma^{-1} \Delta S}
\label{distance}
\end{equation}
where $S=diag([\mathbf{w},\mathbf{w}])$ is the diagonal matrix representing the re-weighting due to the power-spectrum, repeated twice to cover the 22 elements, and $\Delta$ is the difference between the two FRFs expanded to 22 components.
This approach does not require model identification because it is performed on the basis of the data. The comparison can be performed between the tested robot and the average of the groups of humans (healthy or with defined deficient conditions) or between two single samples to quantify how much two robots differ from each other. The score of human-likeness is obtained comparing the sample with the mean of the human sample set $\mathbf{\mu}$ so that 
\begin{equation}
	\Delta=H-\mathbf{\mu}
	\label{Delta}
\end{equation}.
The parameters $\mathbf{\mu}$ and $\Sigma$ defining the score are given in Fig. \ref{fig:Parameters}.
the score in eq. \ref{distance} resembles a Mahalanobis distance 
\begin{equation}
D=\sqrt{\Delta^T \Sigma^{-1} \Delta}
\end{equation},
 but with the addition of weights. Assuming a joint normal distribution for the weighted FRFs, $\mathcal{N}(\mu,\Sigma)$, the Mahalanobis distance defines probability density, where a smaller distance associated with higher probability density.

\section{MATERIALS AND METHODS}
\subsection{Humanoid Robot}
The test is performed with the humanoid robot \textit{Lucy}, a custom platform designed to simulate human posture control\cite{lucyweb}. Lucy Posturob has 14 degrees of freedom controlling posture in the frontal and the sagittal plane. The anthropometric parameters of the robot are shown in Table \ref{anth}

\begin{table}[htb]
\centering
\begin{tabular}{lll}
Body mass                           & 15.30 & kg \\
Upper body mass (including pelvis)  & 9.50  & kg \\
Thighs mass                         & 2.80  & kg \\
Shanks mass                         & 3.00  & kg \\
Total height (from ankle joints)    & 1.52  & m  \\
Body COM height (from ankle joints) & 0.68  & m  \\
Shank COM height                    & 0.32  & m  \\
Thigh COM height                    & 0.31  & m 

\end{tabular}
\caption{Robot anthropometrics.}
\label{anth}
\end{table}
 
\subsection{Control System}
\begin{figure}[htbp]
	\centering
		\includegraphics[width=1.00\textwidth]{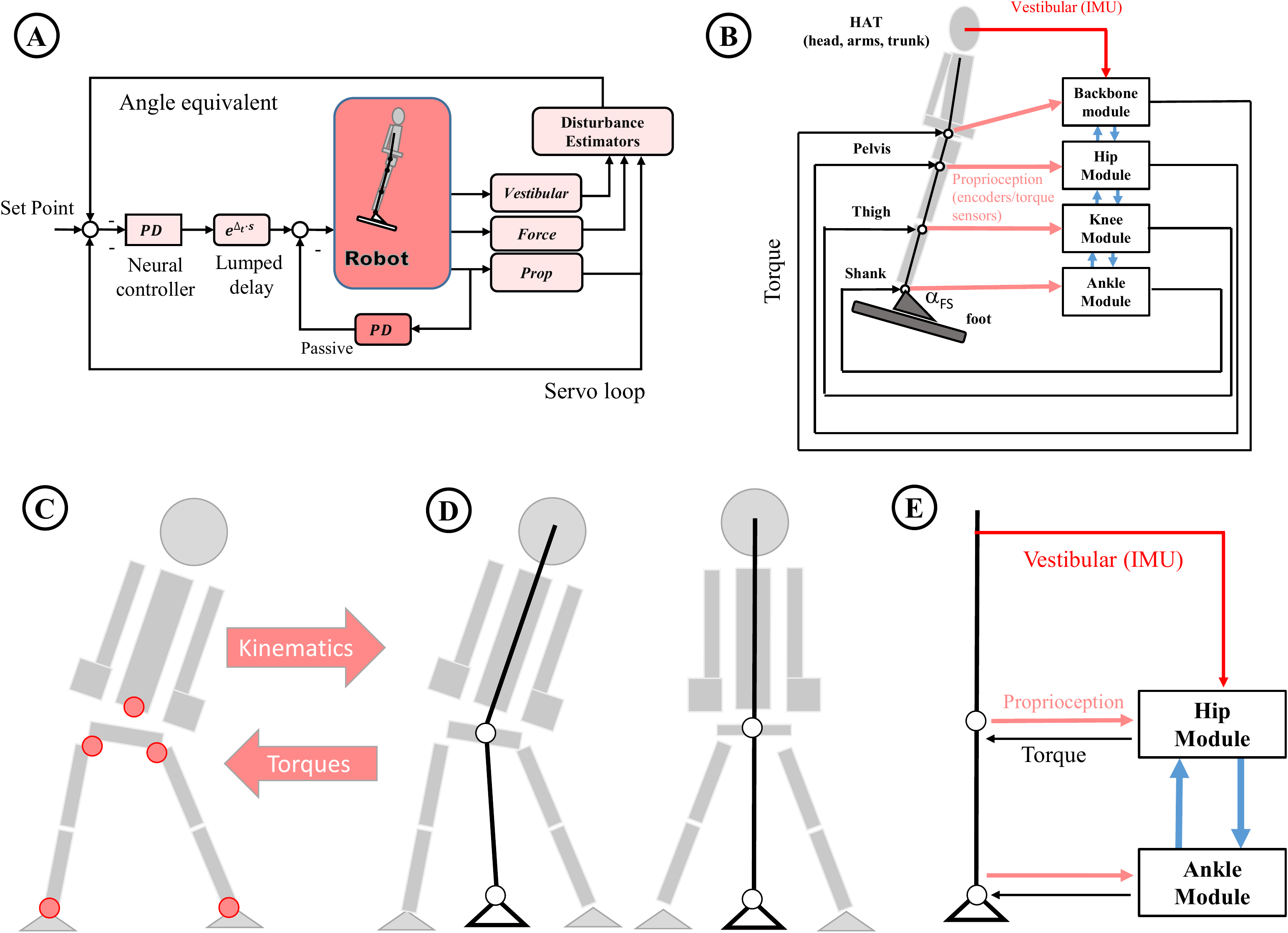}
	\caption{The DEC control system. (A) The servo loop concept showing the compensation of external disturbances. The control scheme is presented for the single inverted pendulum (SIP) case, the generalization to more DOFs is produced with modules controlling each joint as a SIP. 
	(B) The modular contol system in the sagittal plane.
	(C) The modelling in the frontal plane. With fixed feet position and neglecting knee bend, legs, pelvis and the support surface represent a four bar linkage with one DOF, an additional DOF is represented by the link between the hip and the trunk. This is mapped to a DIP model similar to the one used in the description of human balance from \cite{goodworth2010influence}. The torque feedback is computed on the DIP model and then mapped back to the actuators. 
	(D) Two example positions with the associated DIP kinematics.
	(E) The control scheme for the frontal plane with two modules.
	The two degrees of freedom represented by the rotation of the legs around the vertical axis are controlled with a PD feedback (simulating a passive DOF).}
	\label{fig:DEC}
\end{figure}

The robot is controlled using the Disturbance Estimation and Compensation model, DEC\cite{lippi2016human,lippi2017human}. The DEC model	was proposed as a model of human posture control in steady-state \cite{Mergner2003}. The DEC exploits human-like sensor-fusion \cite{Hettich2013,hettich2015human,G.Hettich2014} to reconstruct external disturbances having an impact on body posture and balance, i.e. gravity, support surface tilt, support surface acceleration, external push. The reconstructed disturbances are compensated using a servo controller implemented as a PD. The resulting control law for a joint can be expressed in Laplace domain as
\begin{equation}
T_{joint}= (K_p+sK_d) e^{\Delta_t s} \left( -\epsilon + \alpha_{grav} +\alpha_{trans} +\alpha_{push}  \right) (K_p^{pass}+sK_d^{pass}) \alpha_{joint}
\end{equation}
Where $T_{joint}$ is the torque applied at a joint, $\alpha_{grav}$, $\alpha_{trans}$, and$\alpha_{push}$ are angle equivalent of the external disturbances, i.e. the required torque divided by the proportional gain $K_p$\footnote{In the ideal formulation proposed in \cite{Mergner2010} the controller gain $K_p \cong m \cdot g \cdot h$, where $m$ is the body (or segment) mass, $g$ the gravity acceleration, and $h$ the height of the COM, is equal or slightly smaller than the one required to compensate the gravity torque (that is, with small-angle approximation, $T_g=m \cdot g \cdot h \cdot \alpha_{bs}$).Because of this, the angle equivalent of the gravity torque is the body sway $\alpha_{bs}$. In general additional gains are imposed on the disturbance compensations in order to reproduce human behavior (that does not compensate disturbances perfectly) or to achieve the desired dynamic response on a robot. Notice that the compensating torque is not exactly equal to the estimated disturbance because of the derivative component of the servo loop $sK_d$.}. The support surface tilt is compensated by identifying the support surface tilt and using it in the definition of the controlled error $\epsilon$. The specific definition of $\epsilon$ may change depending on which variable is controlled (e.g. joint angle or COM sway of the above segments).
 The model is nonlinear because it includes sensory thresholds \cite{icinco20mol} and the disturbance estimators are non-linear functions. The control loop includes a lumped delay (one for each joint). Specifically the threshold non-linearity is formulated as:
\begin{equation}
	\hat{\alpha}_{FS}=\int_0^t f_{\theta} \left( \dot{\bar{\alpha}}_{FS} \right)
	\label{fs}
\end{equation}
where $\dot{\bar{\alpha}}_{FS}$ is the support surface rotation velocity (FS = foot-in-space)reconstructed from the vestibular and the proprioceptive input $\hat{\alpha}_{FS}$ is the reconstructed signal used in the servo loop (Fig. \ref{fig:DEC} A).   The function $f_{\theta}$ is the dead-band threshold:
\begin{equation}
	f_{\theta}(\alpha) = \left\{ 	
	\begin{array}{llc}
   		\alpha + \theta & if & \alpha< -\theta \\
			0 & if & -\theta< \alpha< \theta \\
			\alpha - \theta & if & \alpha> \theta \\
 	\end{array}
	\right.
\end{equation}

The formulation of the DEC model leads to a modular control system with one control module for each degree of freedom, based on a servo loop controlled with a PD. The servo loop concept is established in neurological models \cite{wiener1948cybernetics,merton1953speculations}. The modularity, in this specific case, was proposed as a generalization of the two DOF model presented in \cite{G.Hettich2014}. Interestingly the resulting architecture with low-level servo loops resembles state-of-the-art humanoid control systems, e.g. \cite{ishihara2021computationally,hyon2009integration}. The modular control scheme is shown in Fig. \ref{fig:DEC} B.
The control parameters are reported in Table \ref{parametersDEC}
\begin{table}[htbp]
\centering
\begin{tabular}{llll}
\multicolumn{1}{c}{\textbf{Symbol}} & \multicolumn{1}{c}{\textbf{Quantity}}   & \multicolumn{2}{c}{\textbf{Value}} \\ \hline
\textbf{ANKLE}                      &                                         &                 &                  \\
\textit{P}                          & proportional gain                       & 119.57          & Nm/rad           \\
\textit{D}                          & derivative gain                         & 11.95           & Nms/rad          \\
\textit{$G_{ext}$}                       & external torque gain                    & 0.5             &                  \\
\textit{$\omega_{ext}$}                       & External torque filter cutoff frequency & 5               & rad/s            \\
\textit{G}                          & loop gain (multiplies everything)       & 1.2             &                  \\
\textbf{KNEE}                       &                                         &                 &                  \\
\textit{P}                          & proportional gain                       & 55.72           & Nm/rad           \\
\textit{D}                          & derivative gain                         & 0.4458          & Nms/rad          \\
\textit{$G_{ext}$}                       & external torque gain                    & 0.5             &                  \\
\textit{$\omega_{ext}$}                       & External torque filter cutoff frequency & 5               & rad/s            \\
\textbf{HIP}                        &                                         &                 &                  \\
\textit{P}                          & proportional gain                       & 22.71           & Nm/rad           \\
\textit{D}                          & derivative gain                         & 5.67            & Nms/rad          \\
\textit{$G_{ext}$}                       & external torque gain                    & 0.5             &                  \\
\textit{$\omega_{ext}$}                       & External torque filter cutoff frequency & 5               & rad/s    \\       
\textbf{PELVIS}                        &                                         &                 &                  \\
\textit{P}                          & proportional gain                       & 10.59           & Nm/rad           \\
\textit{D}                          & derivative gain                         & 0.07            & Nms/rad          \\
\textit{$G_{ext}$}                       & external torque gain                    & 0             &                  \\
\textit{$\omega_{ext}$}                       & External torque filter cutoff frequency & N.A.               & rad/s

\end{tabular}
\caption{Control parameters for the sagittal plane from\cite{lippi2020challenge}. The $\omega_{ext}$ is the cutoff frequency of a Butterworth filter applied to the external contact torque (push) feedback. A gain G of 1.2 was applied to the output of the PD controller, leading to a slightly overcompensation of the external disturbances. in the here presented experiments this will be modified in order to obtain different responses.}
\label{parametersDEC}
\end{table}

\subsection{Test Protocol}
The robot Lucy has been tested five time with different control system configurations. Specifically the implementation of the DEC from \cite{lippi2020challenge} was used as base and tested, modified version consisted in one without derivative control of the ankle joints and one with twice the derivative gain. Two further trials have been performed with a version of the controller with a lower loop gain (multiplying the output of the PD controller). In one case the gain was 1 and in the other 0.8, versus the 1.2 from \cite{lippi2020challenge}. The body sway used to compute the human likeness measure was computed using the internal sensors of the robot. The signal were synchronized alligning the ideal PRTS samples with the profile of the foot-in-space tilt $\alpha_{FS}$ computed with internal sensors. The alignment was performed on the basis of maximum auto-correlation.

\section{RESULTS}

The results obtained, ordered by score are:
\begin{table}[h!]
\begin{center}
\begin{tabular}{p{0.4\textwidth}p{0.2\textwidth}p{0.2\textwidth}}
{Control} &  \textbf{Score} &   \textbf{CDF}  \\ \cline{1-3}
{No D}                                       &  2.9798         &   92.1053 \%  \\
{Lower gain (1)}                             &  3.1297         &   92.9825 \%  \\
{Lowest gain (0.8)}                          &  3.3708         &   95.6140 \%  \\
{Standard DEC\cite{lippi2020challenge}}      &  3.3823         &   95.6140 \%  \\
{Double D}                                   &  3.5646         &   95.6140 \%  \\
\end{tabular}
\caption{Scores obtained with different control configurations.}
\label{tabres}
\end{center}
\end{table}
 
\begin{figure}[htb!]
	\centering
		\includegraphics[width=1.00\textwidth]{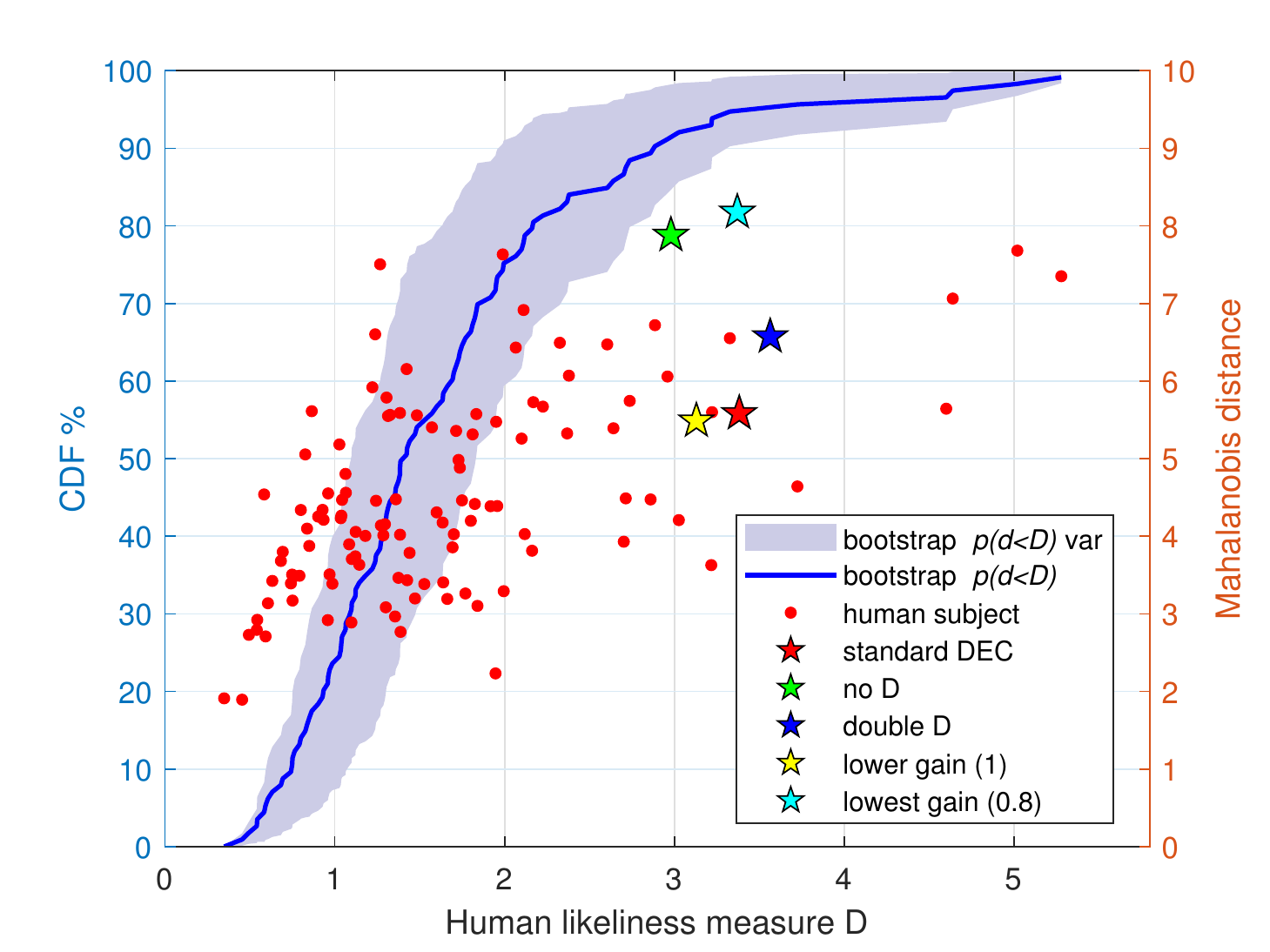}
	\caption{Mahalanobis distance for the samples in the data-set (red dots) as a function of the metric from eq. \ref{distance} and the bootstrap estimated cumulative distribution of the metric (blue line) and its variance (light blue bands)	. The human-likeness score is smaller than the Mahalanobis distance because if the coefficients in Fig. \ref{fig:Parameters} D that are $\leq1$, and in general there is a spread of the Mahalanobis distances of samples associated with the same D because the weighting almost removes the variation due to high-frequency components. The stars represent the results obtained with different configurations.}
	\label{fig:Results}
\end{figure}

The values in the third column give estimates of the cumulative distribution function (CDF) of the data-set at the given score. It is computed by counting the fraction of samples with a score smaller than the one produced in the specific simulation (due to the discrete nature of such measure, two results with a slightly similar score may have the same CDF). Fig. \ref{fig:Results} shows the obtained scores together with human samples for reference. The  Mahalanobis distance is also plotted as function of the score.
\begin{figure}[htb!]
	\centering
		\includegraphics[width=1.00\textwidth]{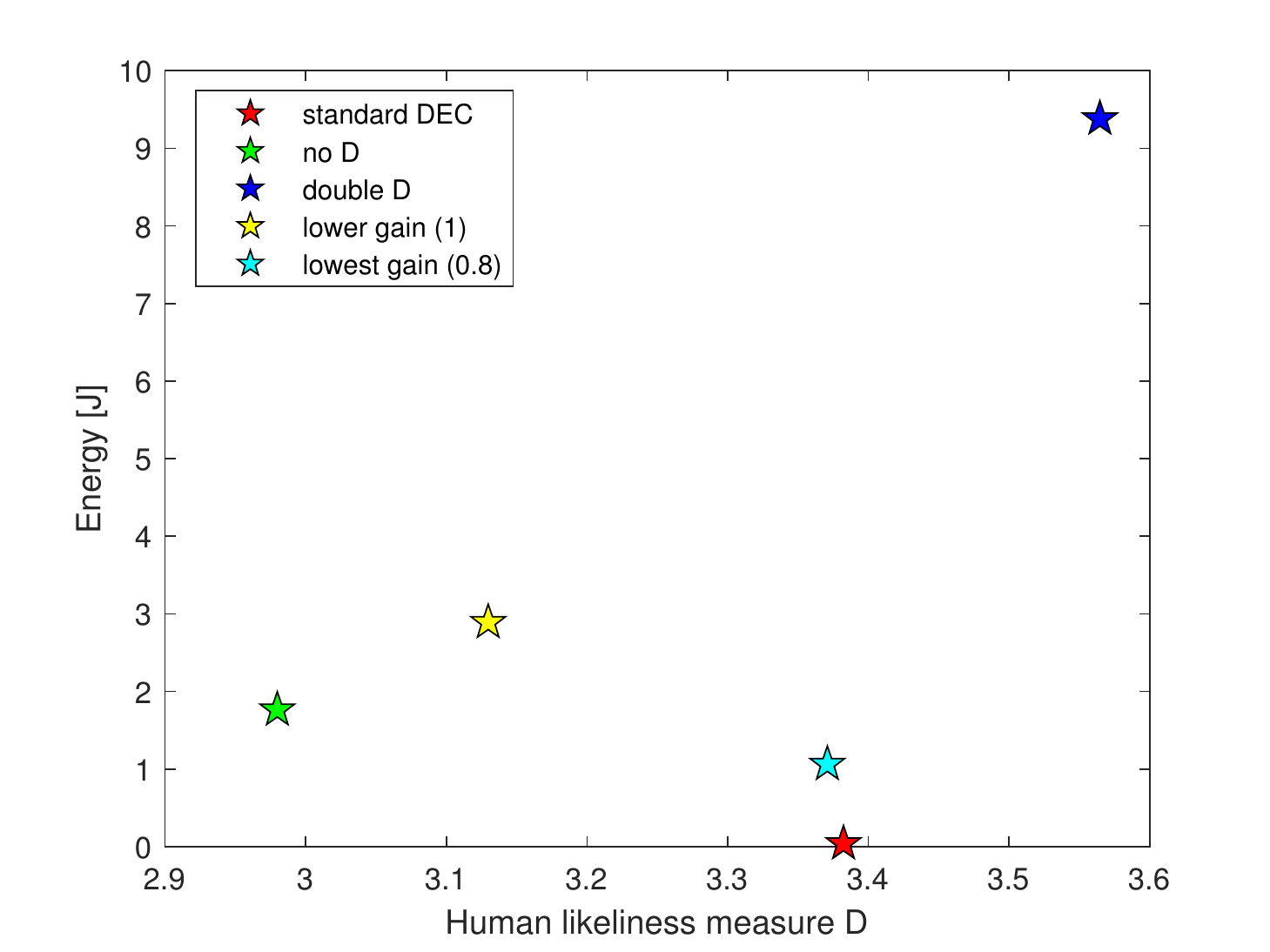}
	\caption{Energy produced by the actuators of the ankles, computed indirectly on the basis of torque and joint angle profiles.}
	\label{fig:Energy}
\end{figure}

\section{Discussion}
The results in Table \ref{tabres} show how the best score was obtained without derivative control on the ankle while the largest score was the one with twice the derivative gain. Loop gains smaller than the base model from \cite{lippi2020challenge} led to better scores although the lowest gain tested (0.8) led to a score that was similar to the one of the original control. Interestingly the result confirms the idea that human like behavior is in general relaxed (i.e. slightly under-compensation of disturbances). The larger Mahlanobis distance exhibited by the controller with the \textit{lowest gain (0.8)} and with \textit{no D} is due to oscillations produced at frequencies that are cut by the weights in eq. \ref{weightsfactors}. Such oscillations are induced by the lack of sufficient damping. One may ask, for example, if the  
human like behavior is advantageous in terms of energy. An indirect estimation of the energy used by the ankle actuators during the trials is shown in Fig. \ref{fig:Energy}. The picture suggests that the most efficient is the original control system (that was carefully tuned to reduce the sway, in contrast to the other that were produced modifying D and G with the sole purpose of creating examples). The lowest gain $G= 0.8$ produces less energy consumption than $G= 1$ but more than the original $G=1.2$. This nonlinear relationship is the result of a trade off between the smaller oscillations produced by higher gains and the smaller torques associated with smaller gains. The Example with double derivative gain is the less human like and the less efficient. 

\section{Conclusions}
This work proposed a score for the evaluation of a human-likeness in humanoid posture control experiments. The experimental set up and the analysis pipeline has been described. The examples shown how different combinations of parameters lead to different scores, suggesting that low loop gain results in more human like FRF profiles. A final consideration on energy consumption by ankle actuators shown that the most human like behavior is not always the most efficient. Hence human-likeness should be associated with other measures to assess the performance of the robot completely. In general human like movements can be an advantage by themselves in human-robot interaction and for assistive devices, while in other contexts it can be sacrificed for efficiency and precision. 

\section{Testbed and code availability}
The presented benchmarking performance indicator is par of the project EUROBENCH \url{https://eurobench2020.eu/}.The EUROBENCH framework includes several tests and performance indicators provided with the aim of testing the performance of robots at any stage of development.

The code will be updated in the docker hub: \\ \url{https://hub.docker.com/repository/docker/eurobenchtest/pi\_comtest}. 

 In Linux, code can be run without being installed locally, as indicated on the \textsl{readme} page: \url{https://github.com/eurobench/pi\_comtest#docker-based-code-access}

\section{Acknowledgment}
\begin{tabular}{l l}
\begin{minipage}{0.1\textwidth}

\includegraphics[width=0.8\textwidth]{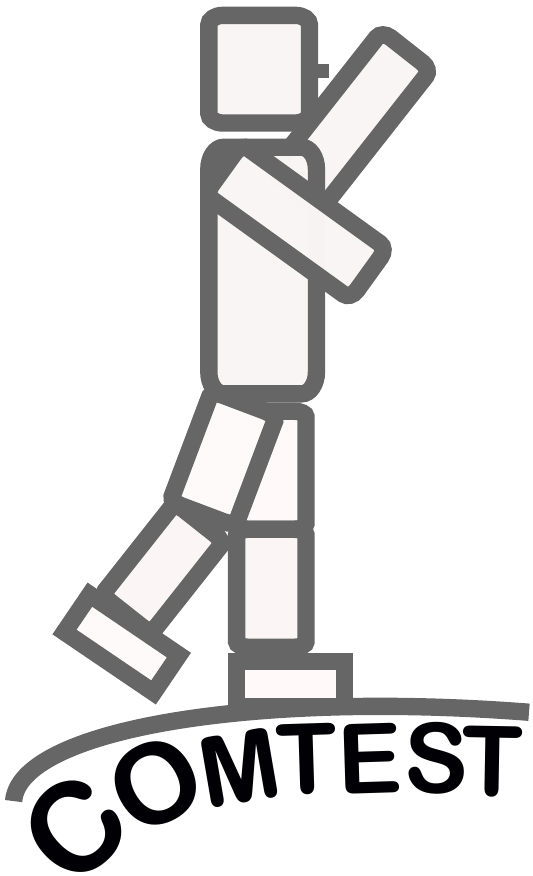}

\end{minipage}
&
\begin{minipage}{0.9\textwidth}
\noindent This work is supported by the project COMTEST, a sub-project of EUROBENCH (European Robotic Framework for Bipedal Locomotion Benchmarking, www.eurobench2020.eu) funded by H2020 Topic ICT 27-2017 under grant agreement number 779963.
\end{minipage}
\end{tabular}

 \bibliographystyle{splncs04}
 \bibliography{example}

\end{document}